\documentclass[a4paper]{IEEEtran}  
\usepackage{caption}

\IEEEoverridecommandlockouts                              

\usepackage{graphicx} 
\usepackage{amsfonts}
\usepackage{amsmath,bm}
\usepackage{caption}

\usepackage{amssymb}
\usepackage{subcaption}
\usepackage{enumerate}
\usepackage{algorithm}
\usepackage{hyperref}

\graphicspath{ {./images/} }
\usepackage[english]{babel}
\newtheorem{theorem}{Theorem}

\author{Share\LaTeX}
\title{\LARGE \bf Non-Normalized Shared-Constraint Dynamic Games for Human–Robot Collaboration with Asymmetric Responsibility}

\author{Mark Pustilnik$^{1}$, Francesco Borrelli$^{1}$
\thanks{$^{1}$University of California at Berkeley, USA \quad {\tt\small \{pkmark, fborrelli\}@berkeley.edu}}%
}

\date{November 2026}

\begin{document}

\maketitle
\thispagestyle{empty}
\pagestyle{empty}

\begin{abstract}
This paper proposes a dynamic game formulation for cooperative human–robot navigation in shared workspaces with obstacles, where the human and robot jointly satisfy shared safety constraints while pursuing a common task. A key contribution is the introduction of a \emph{non-normalized equilibrium} structure for the shared constraints. This structure allows the two agents to contribute different levels of ``effort'' towards enforcing safety requirements such as collision avoidance and inter-players spacing. We embed this non-normalized equilibrium into a receding-horizon optimal control scheme. 
\end{abstract}

\section{Introduction}
The collaborative task problem between two or more players can be formulated with different techniques. Dynamic games is emerging as a prominent and very natural tool that can overcome the shortages of other techniques. If the players are two humans weighted equally, the solution will have different characteristics relative to a human-robot interaction where the robot abilities and effort could be very different. In many scenarios, humans move naturally with minimal attention to constraints, while robots must take on most of the responsibility for enforcing safety boundaries such as collision avoidance or required proximity limits. Classical control approaches typically enforce constraints centrally or assume symmetric responsibility, which does not reflect the inherent asymmetry of human–robot interaction (HRI).

In this paper, we propose a \emph{dynamic game} formulation in which a human and a robot jointly satisfy safety constraints while pursuing a common task. This approach includes the definition of a Generalized Nash Equilibrium Problem (GNEP) and producing a solution called Generalized Nash Equilibrium (GNE). The main novelty of this paper lies in the use of a \emph{non-normalized GNE} to encode asymmetric responsibility for maintaining safety. Unlike normalized GNE, which enforce identical shared Lagrange multipliers, non-normalized GNE allow each agent to contribute different magnitudes to the shared constraint enforcement. This enables the model to represent realistic HRI behavior, where the human does not enforce constraints strongly and the robot compensates to maintain safety and help the human complete the task with minimal effort. The dynamic game framework is very natural in describing the interaction between agents in a competitive and cooperative scenarios, and potentially have many advantages relative to other methods - advantages that dramatically increase when considering non-normalized GNE solutions. While centralized ore decentralized MPC can only express asymmetry through heuristic cost shaping, the non-normalized GNE framework enables a structurally different and mathematically explicit control of how responsibility for shared safety constraints is distributed between agents.

We embed this equilibrium into a receding-horizon optimal control scheme described by Mixed Complementary Problem (MCP) formulation and solve numerically. The resulting architecture retains interpret-ability and modularity while supporting real-time closed loop computation by methods like, imitation learning. The model is evaluated through simulation (Section~\ref{sec:simresults}), illustrating how varying the relative responsibility between the players, and reallocates enforcement effort and reshapes cooperative trajectories. This work follows the work done in \cite{pustilnik2025generalized}.

In this paper we consider the problem of moving a long payload from some initial position to a desired position on a 2D plane. Each player carrying the payload at one end. Since the payload length is fixed, the players needs to make sure the distance between them is also fixed along the path. Each player has its own state and input constraints. In addition, obstacles may be present, such that the players and the payload can't cross the obstacle.

\section{Related Work}
Dynamic games with shared constraints, also known as generalized Nash equilibrium problems (GNEPs), have been extensively studied in mathematical programming and economic optimization. Foundational work includes Rosen’s characterization of Nash equilibria for concave games~\cite{Rosen1965}, and later extensions to shared constraints and complementarity conditions~\cite{FacchineiKanzow2010,Dreves2011}. Non-normalized equilibria, where players’ Lagrange multipliers for shared constraints are not constrained to be identical, have been discussed in theory but remain largely unexplored in robotics and control.

In robotics, MPC methods for multi-agent coordination typically enforce constraints centrally~\cite{Faulwasser2017} or rely on hierarchical decomposition. Game-theoretic formulations have been explored for competitive settings, but cooperative shared-constraint games remain rare, especially in HRI. The notion of allocating different responsibility weights to human and robot players is new and provides a novel way of modeling realistic cooperative behavior.

In Human-Robot Cooperation (HRC), the challenge is to merge human intention with robotic assistance while maintaining safety. Shared autonomy methods~\cite{Hogan2018SharedAutonomy,Argall2018SurveySharedAutonomy} and collaborative manipulation frameworks~\cite{Mainprice2013} model human intent or generate robot motions that account for human predictions. Learning-based approaches have used demonstrations or probabilistic intent models~\cite{Rozo2016LearningHRC,KhansariZadeh2014}, while safety-focused approaches incorporate control barrier functions or physical interaction models~\cite{Cheng2015,Haddadin2017SafePhysicalHRI}.

There are many Model Predictive Control (MPC) algorithms that can deal with multi agents in competitive or cooperative scenarios. The Distributed MPC (DMPC) is a MPC frameworks that solves a different MPC problem for each player and synchronizes the feasibility of the shared constraints in a higher level algorithm, for example in \cite{dunbar2006distributed} an estimated trajectories are only shared with close neighbors. In Centralized MPC (CMPC) a single global optimization problem is solve including all players and constraints, with a single cost function. In \cite{de2023centralized} a CMPC in formulated to solve a Collaborative Loco-Manipulation with legged robots.

More recent work explicitly examines role allocation or responsibility distribution between human and robot partners. Zheng \emph{et al.} propose a trust-driven framework for role adaptation during transport tasks~\cite{zheng2022safe}, while Musi\'c and Hirche formulate shared control as a differential game for haptic collaboration~\cite{music2020haptic}. Our work differs from these by introducing a \emph{non-normalized shared-constraint equilibrium}, wherein asymmetric responsibility emerges directly from differently weighted multipliers in the agents' Lagrangians.

To our knowledge, no prior HRI or MPC-based approach explicitly models asymmetric responsibility for shared safety constraints using non-normalized GNEPs or implements such equilibria in a real-time receding-horizon architecture.

\section{Problem Formulation}

We consider a two-player collaborative motion-planning problem involving a
human (player~1) and a robot (player~2) operating in a planar workspace.
Each agent \(i \in \{1,2\}\) has state variables (position and velocity)
\begin{equation} 
    x_i(k) \in \mathbb{R}^2, \qquad v_i(k) \in \mathbb{R}^2,
\end{equation} 
and control input (acceleration)
\begin{equation} 
    a_i(k) \in \mathbb{R}^2,
\end{equation} 
for discrete time indices \(k = 0,\dots,N\). The horizon length is \(N\) with sampling period \(\Delta t\).

Let the decision vector for player \(i\) be
\begin{equation} 
\begin{aligned}
    z_i = \{\{x_i(k),v_i(k)\}^N_{k=0},\{a_i(k)\}_{k=0}^{N-1}\}
\end{aligned}
\end{equation}

Both players evolve according to a discrete-time double-integrator model:
\begin{equation} 
\begin{aligned}
    x_i(k+1) &= x_i(k) + \Delta t\, v_i(k) + \tfrac{1}{2} \Delta t^{2} a_i(k),\\[2mm]
    v_i(k+1) &= v_i(k) + \Delta t\, a_i(k)
\end{aligned}
\label{eq:dyn}
\end{equation}
Each agent is subject to private velocity and acceleration constraints:
\begin{align}
    \|v_i(k)\|^{2} \le v_{i,\max}^{2}, \qquad
    \|a_i(k)\|^{2} \le a_{i,\max}^{2}.
\end{align}
There are 2 main shared constraints types considered. The first represents that situation where the agents jointly manipulate a payload whose endpoints coincide with their positions. The nominal distance between the agents is \(d > 0\). The payload imposes a rigid-length constraint
\begin{equation}
    \|x_1(k) - x_2(k)\|^{2} = d^{2}.
\end{equation}
To incorporate this relation into a generalized Nash equilibrium framework, we relax the equality into a set of inequality constraints using a small tolerance \(\epsilon > 0\). Meaning for every time step in the prediction horizon the following inequality has to take effect:

\begin{equation} 
\begin{aligned}
    (d - \epsilon)^{2} \le \|x_1(k) - x_2(k)\|^{2} \le (d + \epsilon)^{2}
    \label{eq:dist-band}
\end{aligned}
\end{equation}

The second shared constraint type is obstacle avoidance. Let an obstacle be represented by its center \(p_{\mathrm{obs}} \in \mathbb{R}^{2}\) and radius \(r_{\mathrm{obs}}>0\). Since the payload occupies the entire line segment between the two players, avoidance must be enforced for all points along this segment.

We discretize the segment using interpolation coefficients
\begin{equation} 
    c_f \in \left\{ 0,\;\tfrac{1}{n_p},\; \dots,\;\tfrac{n_p-1}{n_p},\;1 \right\}
\end{equation} 
and define the corresponding intermediate points
\begin{equation} 
    x_{c_f}(k) \;=\; c_f\, x_1(k) + (1-c_f)\, x_2(k).
\end{equation} 
Each of these points must remain outside every obstacle:
\begin{equation} 
\begin{aligned}
    \|x_{c_f}(k) - p_{\mathrm{obs}}\|^{2} \ge r_{\mathrm{obs}}^{2},    \qquad \forall k, ~\forall c_f, ~\forall \text{obstacle}
    \label{eq:obs-avoid}
\end{aligned}
\end{equation} 

Constraints \eqref{eq:dist-band} and \eqref{eq:obs-avoid} couple the decisions of the two players and therefore form the set of \emph{shared constraints} of the game.

Each agent is assigned a quadratic objective that penalizes deviation from its terminal target and control effort. Let \(x_{1,f}, x_{2,f} \in \mathbb{R}^{2}\) denote the desired final positions. For player $i \in \{1,2\}$, the cost is defined by:

\begin{equation} 
\begin{aligned}
    J_i = \sum_{k=1}^{N}  &\|x_i(k) - x_{i,f}\|^2 + \lambda_a \sum_{k=0}^{N-1}\|a_i(k)\|^2
\end{aligned}
\end{equation}

To adapt the dynamic game formulation into a real-world practical solution, extreme situations have to be taken into account. Whether because unsynchronized motion of the players, noisy measured states, or even extreme scenarios, slack variables are introduced into the formulation to help numerical convergence. The slack variables vector $s \in \mathcal{R}^N$ is added to the robot's decision variables ($i=2$):
\begin{equation} 
\begin{aligned}
    \hat{z}_2 = \{\{x_i(k),v_i(k)\}^N_{k=0},\{a_i(k)\}_{k=0}^{N-1},\{s(k)\}_{k=0}^{N}\}
\end{aligned}
\end{equation}
The slack variables are added to the robot's cost function to minimize there use. The human's Cost function stays unchanged:
\begin{equation} 
\begin{aligned}
    \hat{J}_1 &= J_1 = \sum_{k=1}^{N}  \|x_1(k) - x_{1,f}\|^2 + \lambda_a \sum_{k=0}^{N-1}\|a_1(k)\|^2 \\
    \hat{J}_2 &= \sum_{k=1}^{N}  \|x_2(k) - x_{2,f}\|^2 +... \\& ~~~~~~~~~~~~~ ...+\lambda_a \sum_{k=0}^{N-1}\|a_2(k)\|^2+\lambda_s \sum_{k=0}^{N}s(k)^2
\end{aligned}
\end{equation}
where $\lambda_s$ is the slack variables penalty weight.
The Shared constraint that enforces distance between players becomes:
\begin{equation} 
\begin{aligned}
    (d - \epsilon)^{2}- s(k) \le \|x_1(k) - x_2(k)\|^{2} \le (d + \epsilon)^{2} + s(k)
    \label{eq:dist-band}
\end{aligned}
\end{equation}

The resulting dynamic game is the following GNEP: for each \(i \in \{1,2\}\),
\begin{equation}
\label{eq:gnep}
\begin{aligned}
    &\min_{z_i} \quad  \hat{J}_i(z_1,z_2)
    \\[1mm]
    &\text{s.t.}\quad \\
    & x_i(0)=x_{i,0}, \qquad v_i(0)=v_{i,0},
    \\
    & x_i(k+1) = x_i(k) + \Delta t\, v_i(k)
        + \tfrac{1}{2}\Delta t^2 a_i(k),
    \\
    & v_i(k+1) = v_i(k) + \Delta t\, a_i(k),
    \quad k = 0,\dots,N-1,
    \\
    & \|v_i(k)\|^{2} \le v_{i,\max}^{2},
    \qquad k = 1,\dots,N,
    \\
    & \|a_i(k)\|^{2} \le a_{i,\max}^{2},
    \qquad k = 1,\dots,N-1,
    \\
    & (d - \epsilon)^{2} -s(k)
        \le \|x_1(k) - x_2(k)\|^{2}
        \le (d + \epsilon)^{2}+s(k),
    \\
    & \|x_{c_f}(k) - p_{\mathrm{obs}}\|^{2}
        \ge r_{\mathrm{obs}}^{2},
        \qquad 
        \forall k,\;\forall c_f,\;\forall \text{obstacles}.
\end{aligned}
\end{equation}

\subsection{Generalized Nash Equilibrium Problem}
The interaction between the human (player~1) and the robot (player~2) is modeled as a \emph{Generalized Nash Equilibrium Problem} (GNEP) due to the presence of shared constraints that couple the decision variables of both players. Unlike standard Nash games, where each player’s feasible set depends solely on their own strategy, here the feasible sets of both agents are jointly restricted by the distance constraints~\eqref{eq:dist-band} and the obstacle-avoidance constraints~\eqref{eq:obs-avoid}. Consequently, each player faces an optimization problem whose feasible region depends explicitly on the other agent’s decision vector.

Let \(Z_i\) denote the feasible set of player \(i\), defined by its private dynamics, input bounds, and the shared constraints that involve both \(z_1\) and \(z_2\). A pair \((z_1^\star, z_2^\star)\) is a \emph{generalized Nash equilibrium} if, for each \(i \in \{1,2\}\),
\begin{equation}
    z_i^\star \in \arg\min_{z_i \in Z_i(z_{-i}^\star)} J_i(z_i, z_{-i}^\star),
\end{equation}
where \(z_{-i}\) denotes the other player's decision vector. At equilibrium, no player can unilaterally modify its state–input trajectory to decrease its own cost while remaining feasible with respect to both its individual constraints and the shared constraints induced by the other player's strategy. The concept of Nash Equilibrium (NE) as solution to a dynamic game is widely used and presents a numerous advantages over other solution concepts. Usually in GNEP the set of possible solutions is not unique - there could be many possible GNE. The selection of a specific solution is discussed next.

\section{Non-Normalized Vs Normalized GNE}

In generalized Nash equilibrium problems with shared constraints, the coupling between players is captured not only in the primal feasibility conditions but also in the dual variables associated with the shared constraints. The standard notion of equilibrium used in most of the literature is the \emph{normalized generalized Nash equilibrium} (normalized GNE), originally formalized by Rosen \cite{Rosen1965}. In this equilibrium concept, the Lagrange multipliers associated with each shared constraint are required to be equal across all players. This enforces a symmetric treatment of the shared constraints: each player is assigned the same ``responsibility'' in satisfying the constraint, and the dual variables do not encode any asymmetry between agents.

Formally, the Lagrangian of player $i \in \{1,2\}$ is defined by:
\begin{equation} \label{Lagrangian}
    \mathcal{L}_i := J_i(z^i,z^{-i})+ \mu_i^Th_i(z^i) + \lambda_i^T \cdot g_i(z^i) + \sigma_i^T \cdot s(z^i,z^{-i})
\end{equation}
where, $h_i(z_i), ~\mu_i \in \mathbb{R}^{k_i}$ are the dynamics functions and Lagrange multipliers of the equality constraints of player $i$ respectively, $g_i(z^i),~\lambda_i \in \mathbb{R}^{m_i}$ are the private inequality constraints and Lagrange multipliers of the private constraints of player $i$ respectively, and $s(z_i,z_{-i}), ~\sigma_i \in \mathbb{R}^{m_0}$ are shared constraints and the Lagrange multipliers of the shared inequality constraints of player $i$ respectively.
For a point $z^* =\{z_1^*,z_2^*\}$ to be a GNE, the following KKT conditions have to be satisfied for $i \in \{1,2\}$:
\begin{equation} \label{KKT}
\begin{aligned}
    & \nabla_{z_i}\mathcal{L}_i =0, \qquad  & h_i = 0,  \\
    & 0 \leq \lambda_i \perp g_i \leq 0,  \qquad   & 0 \leq \sigma_i \perp s \leq 0,  \\
    & \lambda_i \geq 0, \qquad   & \sigma_i \geq 0,  \\
\end{aligned}
\end{equation}

A normalized GNE satisfies the condition
\begin{equation}
    \sigma_1 = \sigma_2,
\end{equation}
together with the individual optimality conditions of each player. This choice yields a variational equilibrium and is mathematically convenient because it leads to a single aggregated KKT system. However, it implicitly assumes that both agents influence the shared constraints in an identical manner.

In many human-robot collaboration settings, the agents do not share the same level of authority, initiative, responsibility, or physical capability. For example, a human operator may want to minimize their own effort while giving the robot the responsibility to take care of most of the hard work. Such scenarios motivate the use of \emph{non-normalized generalized Nash equilibria}, where the multipliers associated with the shared constraints are no longer required to be equal.

Following the work done in \cite{pustilnik2025generalized}, we can calculate a GNE solution where the human's Lagrange multipliers associated with the shared constraints are different than the robot's. We bring here the main theorem proven in \cite{pustilnik2025generalized} for completeness. Define for every player $i$ a diagonal matrix of weights:
\begin{equation} \label{factors}
\begin{aligned}
    &~~~~~~~~~~~~~~A_i \in \mathcal{D}^+_m, \\
    \mathcal{D}^+_m := \{&D \in \mathbb{R}^{m \times m} \mid D = \text{diag}(w_1, \cdots, w_m), \\ 
    &w_i \in \mathbb{R}_{++} \, \forall i = 1, \cdots, m \}.
\end{aligned}
\end{equation}

\begin{theorem} \label{thm1}
    Given a set of matrices $\{A_i\}_{i=1}^M \in \mathcal{D}^+_{m_i}$ and the tuple $(\bar{x}, \{\bar{\mu}_i\}_{i=1}^M, \{\bar{\lambda}_i\}_{i=1}^M, \bar{\sigma})$ that solves the KKT conditions in (\ref{KKT}), then $\bar{x}$ is a solution to the GNEP if a suitable constraint qualification holds.
\end{theorem}

Theorem \ref{thm1} gives a practical method to calculate a non-normalized GNE solutions of a GNEP. For every choice of the parameter matrices $\{A_i\}_{i=1}^M$, a GNE solution can be calculated, if  exists. 

In general, every single shared constraint can get a different scaling parameter. To reduce the scale of the number of parameters in the HRI investigated here, we define a single parameter $\alpha$ that scales the shared constraints of the human's and robot's Lagrange multipliers :
\begin{equation}
    \sigma_1=\alpha \cdot\sigma, \qquad \sigma_2=(1-\alpha) \cdot \sigma
\end{equation}
where, $\sigma_1$ and $\sigma_2$ are the Lagrange multipliers associated with the shared constraints, and $\sigma$ is the unscaled Lagrange multiplier,
for some responsibility coefficient $\alpha\in (0,1)$. In general we could define a different scaling parameter for every term in the Lagrange multiplier vector associated with the shared constraints $\sigma_i$, but for practical reasons and for numerical interoperability of the problem, we restrict ourselves to a single scalar $\alpha$. The parameter $\alpha$ determines how the load of satisfying the shared constraints is distributed between the two agents. The case $\alpha = 0.5$ recovers the normalized equilibrium, while $\alpha \neq 0.5$ introduces an asymmetric influence in the equilibrium conditions. When $\alpha > 0.5$, player 1 bears a greater share of responsibility in enforcing the shared constraints; conversely, $\alpha < 0.5$ shifts more responsibility to player 2.

The non-normalized framework provides a richer and more flexible set of GNE solutions. It enables modeling situations where one agent is more proactive or more compliant than the other. In dynamic human-robot collaboration, these asymmetries are essential - they allow the equilibrium solution to encode behavioral roles such as leader/follower, dominant/compliant, or assertive/passive interaction patterns, all while remaining within a principled game-theoretic formulation.

In the remainder of this work, we adopt the non-normalized equilibrium concept and show how the shared-constraint multipliers, scaled by the factor $\alpha$, naturally induce asymmetric responsibility allocation that affects the resulting motion trajectories and resulting effort by the players.

Once the parameter $\alpha$ was selected, the dynamic game can be solved. The KKT equations are formulated as a Mixed Complementarity Problem (MCP), and solved using a numerical solver. The dynamic game is solved iteratively at each time step from the initial condition of each player. The first control action is applied and the game is solved again. The dynamic game is solved using the PATH numerical solver \cite{FerrisMunsonPATH}.

\section{Simulation Results} \label{sec:simresults}
We consider a 2-player setup where each player handles an edge of a long payload that needs to be moved from some initial position to a final position. The human player (player 1) has different abilities than the robot player (player 2). The dynamic game prediction horizon is $N=10$ time steps where each time step is $dt=0.1[sec]$. The payload length is $L=3[m]$, and the relaxation parameter $\epsilon=0.05[m]$. This relaxation is very practical since in most practical cases the distance between players is constrained by the payload length and the reach of the players (e.g., human's arms length). This is translated to a different state and position feasible set for each player. Table ~\ref{tab:human_robot_parameters} summarizes the parameters of each player:

\begin{table}[h]
\centering
\begin{tabular}{|c|c|c|c|}
\hline
\textbf{Parameter} & \textbf{Human} & \textbf{Robot} & \textbf{units} \\
\hline
Maximum velocity $v_{i,max}$ & 1.5 & 3 & $[m/s]$ \\
\hline
Maximum Acceleration $a_{i,max}$ & 5 & 25 & $[m/s^2]$ \\
\hline
\end{tabular}
\caption{Comparison between Human and Robot Parameters}
\label{tab:human_robot_parameters}
\end{table}

Next we consider a few different scenarios to see the behavior of each player as a function of the responsibility parameter $\alpha$. The first scenario has no obstacles and both players play according to the GNE solution of the same game. The second scenario is the same as scenario 1, but the human player follows a predefined trajectory, and the robot player plays according to the GNE solution. The third Scenario is similar to scenario 1 in added obstacles. Table \ref{tab:scenarios} summarizes the scenarios characteristics.

\begin{table}[h]
\centering
\resizebox{1.0\linewidth}{!}{%
\begin{tabular}{|c|c|c|c|}
\hline
\textbf{} & \textbf{Scenario 1} & \textbf{Scenario 2} & \textbf{Scenario 3} \\
\hline
Human Initial Position & $[-1.5,6.5]$ & $[-1.5,6.5]$ & $[-1.5,6.5]$ \\
\hline
Human Final Position & $[8.0,2.0]$ & $[8.0,2.0]$ & $[8.0,2.0]$ \\
\hline
Robot Initial Position & $[-3.62, 4.38]$ & $[-3.62, 4.38]$ & $[-3.62, 4.38]$ \\
\hline
Robot Final Position & $[8,5]$ & $[8,5]$ & $[8,5]$ \\
\hline
Human Strategy & Dynamic Game & Dynamic Game & Predefined Path \\
\hline
Robot Strategy & Dynamic Game & Dynamic Game & Dynamic Game \\
\hline
Obstacles & None & Exist & Exist \\
\hline
\end{tabular}
}
\caption{Scenarios Description}
\label{tab:scenarios}
\end{table}
The entire code used to produce the results in this paper is available in \href{https://github.com/pukmark/HRC_Simulation}{Project GitHub \cite{pukmark_hrc_simulation}}

\subsection{Scenario 1}
\begin{figure}[t]
    \centering
    \includegraphics[width=1.0\linewidth]{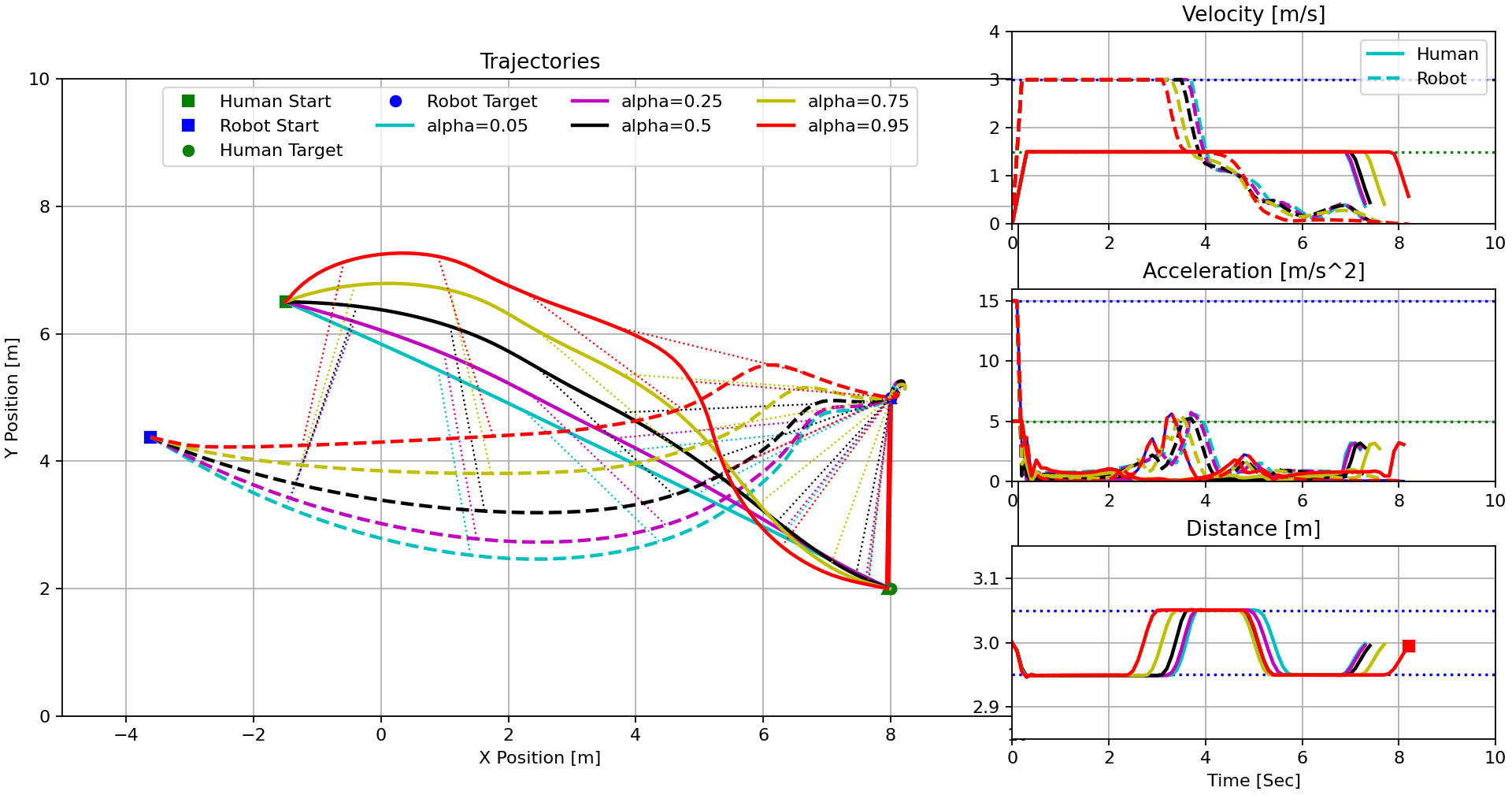}
    \caption{Scenario 1 Trajectories}
    \label{fig:Scenario1}
\end{figure}

In the first scenario, both players play according to the results of the dynamic game. First the scenario is run with $\alpha=0.05$, which corresponds to a situation where the human exerts minimal effort possible to complete the task while the robot reacts to the human's will. This is essentially a leader–follower scenario. Figure \ref{fig:Scenario1} shows the resulting trajectory of both players (\href{https://github.com/pukmark/HRC_Simulation/blob/main/Videos/Scenario_1.mp4}{Video}). The left Subfigure shows the trajectory of each player for different values of $\alpha$. For $\alpha=0.05$, the human is the leader and the robot follows - the human goes straight to its final position and the robot makes a long path toward its final position while not interrupting the human's path. The figure also shows the velocity, acceleration and the distance between the players.
It can be seen that there is a visible difference between the trajectories for different $\alpha$. This scenario demonstrates the influence of the dominance parameter on the trajectories taken by each player. When the human dominates and the robot only responds, the human's effort is minimal, while in the normalized case ($\alpha=0.5$) the effort of both players is more equal.

\subsection{Scenario 2}
\begin{figure}[t]
\centering

\begin{subfigure}{0.9\columnwidth}
\centering
\includegraphics[width=\linewidth]{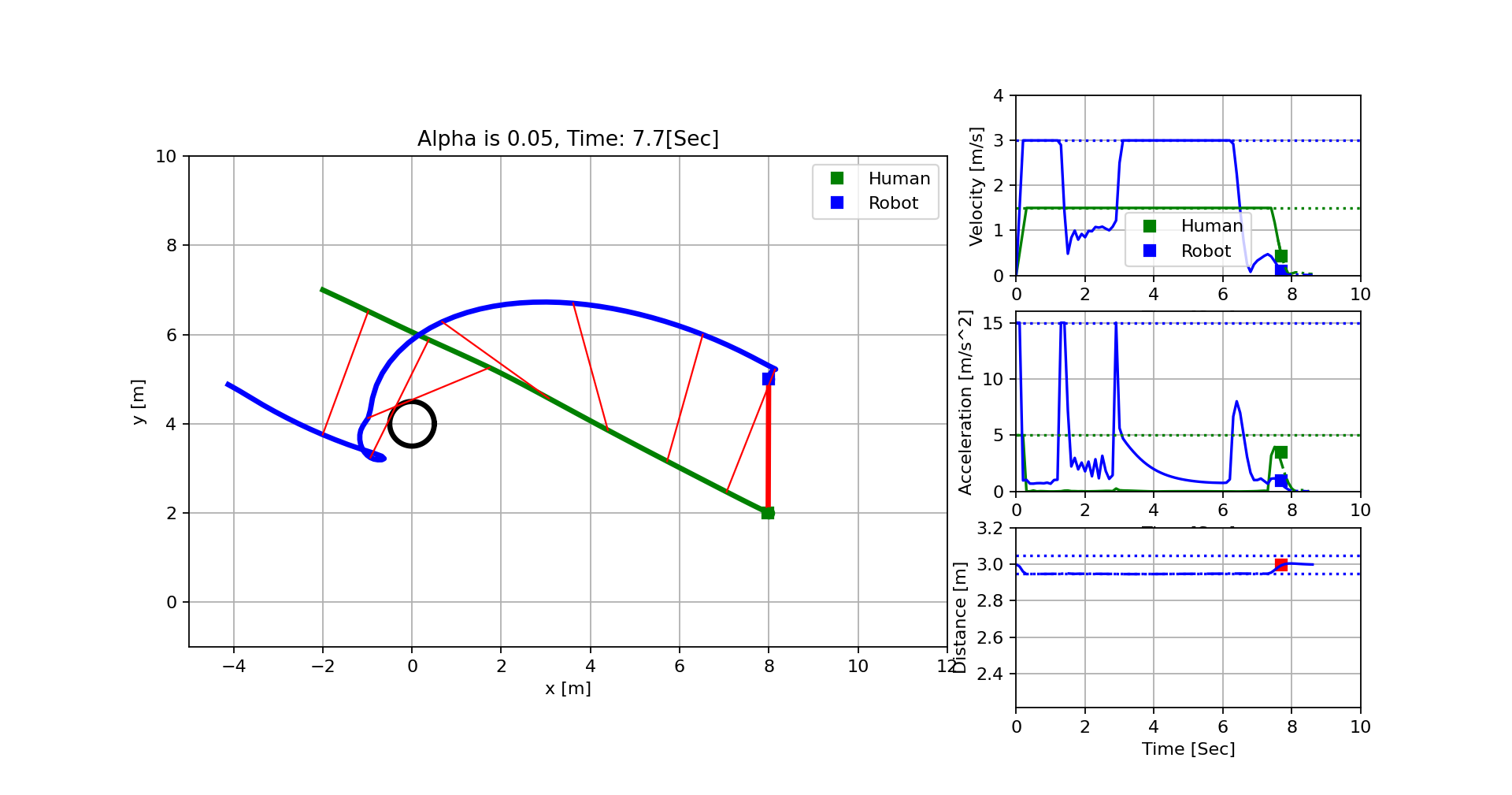}
\caption{Small $\alpha$ (leader-follower behavior)}
\label{fig:scenario2_alpha}
\end{subfigure}

\vspace{1mm}

\begin{subfigure}{0.75\columnwidth}
\centering
\includegraphics[width=\linewidth]{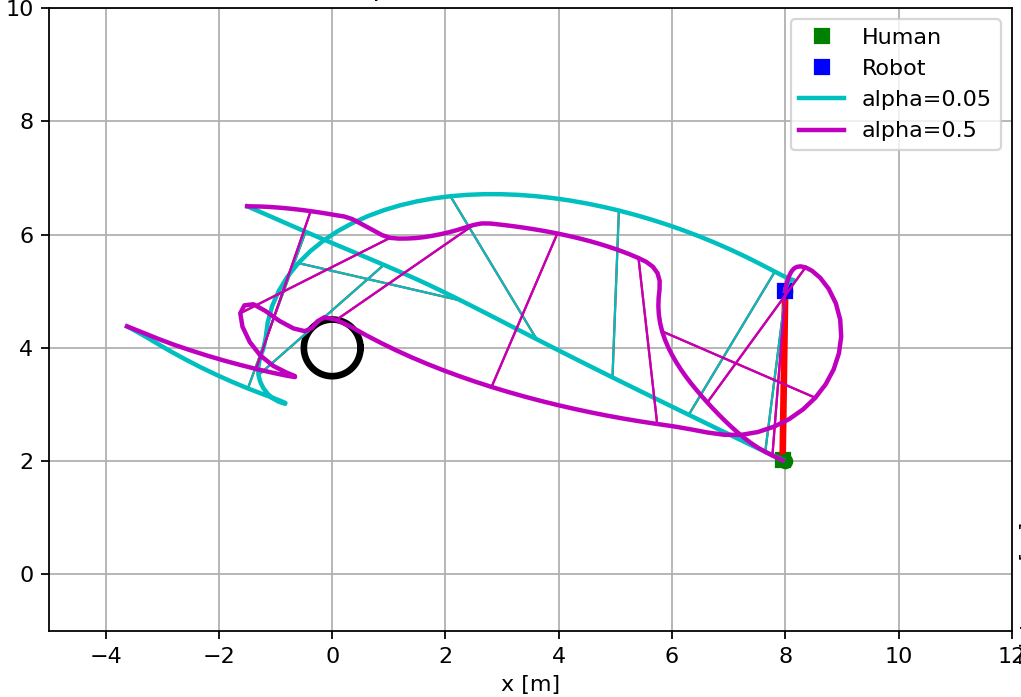}
\caption{leader-follower ($\alpha = 0.05$) Vs Normalized case ($\alpha = 0.5$)}
\label{fig:scenario2_paths}
\end{subfigure}

\caption{Scenario 2 — Trajectories of both players for different values of $\alpha$.}
\label{fig:Scenario2}
\end{figure}

In this scenario, we introduce an obstacle that intersects with the path of the players and the have to respond (\href{https://github.com/pukmark/HRC_Simulation/blob/main/Videos/Scenario_2.mp4}{Video}). Figure \ref{fig:Scenario2} show the trajectory of the players where both players play according to cooperative and synchronized game solution with $\alpha=0.05$, which means the human is the leader and dominant, and the robot tries to respond. Subplot \ref{fig:scenario2_paths} shows the difference between the normalized solution and the leader-follower non-normalized solution. It can be seen clearly that the later is much simpler and more intuitive. The human's path in the normalized solution is non trivial and obviously not something that a human can perform robustly in real-time. In this scenario the robot insists in taking the more aggressive path, even though the resulting trajectory is longer and significantly more complex.

\subsection{Scenario 3}
This scenario demonstrates the robustness of the non-normalized solution relative to the normalized solution in a non-cooperative and unsynchronized scenarios. In this scenario, the human's trajectory is control by following a virtual target that converges to the desired final position, and executed without any feedback from the robot, while the robot plays according to the game solution. In addition, a random normally distributed noise is add to the the human's acceleration command. This simulates the randomness in the motion of humans and its unpredictability:
\begin{equation} 
\begin{aligned}
    a_{1,x}(k) = a^{VT}_x + \mathcal{N}(0,(\frac{a_{1,max}}{5})^2) \\
    a_{1,y}(k) = a^{VT}_y + \mathcal{N}(0,(\frac{a_{1,max}}{5})^2)
\end{aligned}
\end{equation}

\begin{figure}[t]
    \centering
    \includegraphics[width=1.0\linewidth]{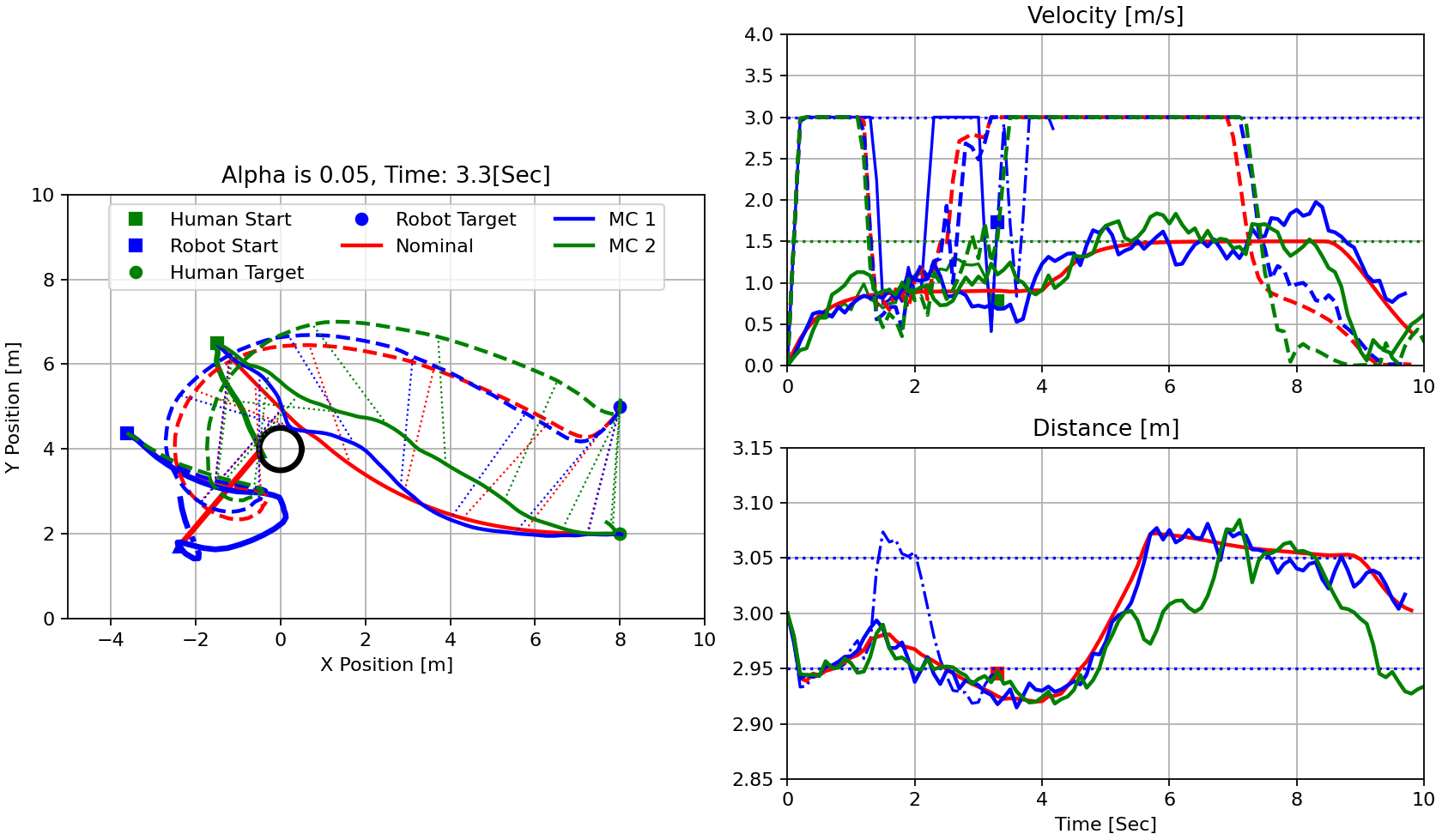}
    \caption{Scenario 3 Monte-Carlo sample Trajectories}
    \label{fig:Scenario3}
\end{figure}

Figure \ref{fig:Scenario3} shows a different realizations of the trajectories of both players, with the nominal trajectory of the human (no noise on acceleration command). 

The idea behind the non-normalized with small $\alpha$ is that when the players come close to a potentially dangerous situation to become infeasible, the robot  will react earlier to resolve the situation relative to the normalized solution in which the robot will assume some response from the human - the robot will not assume the human will act to resolve the situation. 200 Monte-Carlo (MC) were preformed of the scenario for every value of $\alpha$ and preform a statistical analysis on the performance of the robot. The performance metrics are as follows:
\begin{itemize}
    \item Success Percentage - The percentage of scenarios that obstacle avoidance is preserved throughout the scenario
    \item Mean Distance - the average, over Monte Carlo runs, of the time-averaged distance between players
    \item Mean Max Distance - the mean of the maximal distance of each monte-carlo run
    \item Mean effort - the average, over Monte Carlo runs, of the time-averaged acceleration command magnitude of the robot
    \item Std effort - the standard deviation, over Monte Carlo runs, of the time-averaged acceleration command magnitude of the robot
\end{itemize}

Table \ref{tab:mc_resutls} summarizes the results.

\begin{table}[t]
\scriptsize
\centering
\begin{tabular}{c|cccc}
\hline
 & $\alpha = 0.05$ & $\alpha = 0.1$ & $\alpha = 0.3$ & $\alpha = 0.5$ \\
\hline
Success Percentage [\%] & 75[\%] & 73[\%] & 71.5[\%] & 59.5[\%] \\
Mean Distance [m] & 0.049 & 0.050 & 0.061 & 0.097 \\
Mean Max Distance [m] & 0.122 & 0.119 & 0.195 & 0.382 \\
Mean effort [m/$s^2$] & 3.761& 3.766 & 3.977 & 4.072 \\
Std effort [m/$s^2$]& 3.72 & 3.711 & 3.86 & 3.91 \\

\hline
\end{tabular}
\caption{Scenario 3 - MC Results Summery for different $\alpha$ values}
\label{tab:mc_resutls}
\end{table}

It can be seen that, for a lower values of $\alpha$, the performance of the robot in terms of keeping the obstacle avoidance constraints, keeping distance to the human and the control effort is improving. This means, that when the movements is not synchronized between players, the closer the robot behaves like a follower, the better the overall performance. Meaning, when the motion of players is not synchronized, the robot should assume a more "follower" behavior to reduce its control effort and ability to keep shared constraints.

\section{Conclusions}

This paper introduced a dynamic game formulation for cooperative human-robot navigation with shared safety constraints and asymmetric responsibility. By leveraging a non-normalized (GNE), the proposed framework enables explicit and continuous allocation of responsibility between the players through a single scalar parameter $\alpha$. Unlike classical normalized GNE formulations, which enforce identical Lagrange multipliers for shared constraints, the non-normalized structure allows the human and robot to contribute different levels of effort toward maintaining safety, capturing a fundamental asymmetry that naturally arises in human-robot collaboration.

A dynamic game embedded into a receding-horizon Model Predictive Control (MPC) architecture is a natural framework to describe and solve this problem. The resulting closed-loop system preserves modularity and interpretability. The shared-constraint formulation simultaneously enforces payload constraints and obstacle avoidance, ensuring coordinated and physically feasible motion of both players. The framework presented here is a general framework that can work in many scenarios where shared constraints and asymmetric responsibility for safety is relevant and may be important. In addition, in this paper the dominance parameter $\alpha$ stays static throughout the scenario. Future research can be done on how to dynamically change $\alpha$ to get better results - for example, the robot raises a flag that he doesn't has a feasible solution and the human must take more responsibility maintaining safety.

Simulation results across three scenarios demonstrated the practical impact of asymmetric responsibility allocation. In cooperative leader/follower scenarios, small values of $\alpha$ produced intuitive behaviors in which the human follows near-straight paths while the robot assumed most of the constraint-enforcement burden. When obstacles are present, the non-normalized equilibrium generated smoother and more predictable human-robot trajectories than the normalized case. In non-cooperative scenarios with a pre-defined human trajectory, the non-normalized GNE exhibited superior robustness and feasibility, whereas the normalized solution fails more due to delayed robot reaction. These results highlight the advantages of the proposed approach in terms of safety, robustness, and behavioral interpretability.

Future work can focus on extending the framework to learned human intent models using the dynamic game formulation as expert in Reinforcement Learning (RL) methods, uncertainty-aware dynamic game formulations, and multi-agent teams. Experimental validation on physical human-robot environments is also planned, to verify the applicability of this methods in practical scenarios


\bibliographystyle{IEEEtran}
\bibliography{IEEEfull,bibliography.bib}

\end{document}